\documentclass[10pt, a4paper]{article}
\usepackage[final]{lrec-coling2024} 
\usepackage[T1]{fontenc}
\usepackage{graphicx}
\usepackage{tabularx}
\usepackage{soul}
\usepackage{booktabs}
\usepackage{multirow}
\usepackage{amsmath}
\usepackage{rotating}
\usepackage{titlesec}
\usepackage{makecell}
\usepackage{pdflscape}
\usepackage{xcolor}
\usepackage{hyperref}

\usepackage{fancyhdr}
\usepackage[bottom]{footmisc}
\fancypagestyle{firstpagestyle}
{
   \fancyhf{}
   \fancyfoot[C]{\footnotesize This work was published in the proceedings of the 2024 Joint International Conference on Computational Linguistics, Language Resources and Evaluation (LREC-COLING 2024).\\The final version is available at \url{https://aclanthology.org/volumes/2024.lrec-main/}.}
}

\definecolor{darkblue}{rgb}{0, 0, 0.5}
\hypersetup{colorlinks=true, citecolor=darkblue, linkcolor=darkblue, urlcolor=darkblue}
\definecolor{lightgray}{rgb}{0.9,0.9,0.9}

\usepackage{xstring}
\usepackage{color}
\usepackage{appendix}
\usepackage[capitalize]{cleveref}

\AtBeginEnvironment{appendices}{\crefalias{section}{appendix}}

\usepackage{subcaption}
\usepackage{enumitem}
\usepackage{graphicx}
\usepackage{soul}
\usepackage{relsize}
\usepackage{colortbl}
\usepackage{algorithm,algpseudocode}
\usepackage{vwcol}

\newcommand{\ccell}[1]{\multicolumn{1}{c}{#1}}

\makeatletter
\renewcommand\@makefntext[1]{\leftskip=1em\hskip-0.4em\@makefnmark#1}
\makeatother

\sethlcolor{gray!30}

\crefformat{section}{\S#2#1#3} 
\crefformat{subsection}{\S#2#1#3}
\crefformat{subsubsection}{\S#2#1#3}

\newcolumntype{H}[1]{>{\hsize=#1\hsize\arraybackslash}X}
\newcolumntype{T}[1]{>{\hsize=#1\hsize\centering\arraybackslash}X}
\newcolumntype{C}[1]{>{\arraybackslash}p{#1}}

\title{SignBLEU: Automatic Evaluation of \\Multi-channel Sign Language Translation}

\name{Jung-Ho Kim$^\ast$\thanks{$^\ast$Equal contributions.}, Mathew Huerta-Enochian$^\ast$, Changyong Ko, Du Hui Lee$^\dagger$\thanks{$^\dagger$Corresponding author.}}

\address{EQ4ALL\\
         Nonhyeon-ro 76-gil 11, Gangnam-gu, Seoul, Republic of Korea\\
         \{stuartkim, mathew, ericko, scottlee$^\dagger$\}@eq4all.co.kr\\}

\abstract{
Sign languages are multi-channel languages that communicate information through not just the hands (manual signals) but also facial expressions and upper body movements (non-manual signals). However, since automatic sign language translation is usually performed by generating a single sequence of glosses, researchers eschew non-manual and co-occurring manual signals in favor of a simplified list of manual glosses. This can lead to significant information loss and ambiguity. In this paper, we introduce a new task named multi-channel sign language translation (MCSLT) and present a novel metric, \texttt{SignBLEU}, designed to capture multiple signal channels. We validated \texttt{SignBLEU} on a system-level task using three sign language corpora with varied linguistic structures and transcription methodologies and examined its correlation with human judgment through two segment-level tasks. We found that \texttt{SignBLEU} consistently correlates better with human judgment than competing metrics. To facilitate further MCSLT research, we report benchmark scores for the three sign language corpora and release the source code for \texttt{SignBLEU} at \url{https://github.com/eq4all-projects/SignBLEU}.
\\ 
\newline  }

\begin{document}

\maketitleabstract

\thispagestyle{firstpagestyle}

\section{Introduction}
\label{sec:intro}
Sign language translation (SLT) is an emerging field that aims to bridge the gap between the Deaf, hard-of-hearing, and hearing communities. 
With the introduction of neural machine translation, SLT has experienced significant advancements~\cite{camgoz2018neural}, and innovative strategies for generating poses and videos continue to be developed~\cite{stoll2020text2sign,Saunders_2022_CVPR}.

A common approach to text-to-sign translation is to predict glosses, semantic labels for individual signs~\cite{muller23considerations}.
Gloss-based SLT represents signing as a single sequence of gloss tokens, standard sequence-to-sequence modeling techniques can be used, allowing researchers to leverage the capabilities of pre-trained language models~\cite{lee2023leveraging}.
However, by limiting translation to a linear gloss sequence, non-manual expressions that encapsulate additional semantic and morphological aspects of sign language as well as co-occurring manual signals are omitted. Non-manual signals convey important descriptive information, often playing the role of adjectives and adverbs~\cite{crasborn2008frequency,herrmann2013modal}. For example, mouthings can differentiate between identical manual signals~\cite{woll2001he,crasborn2008frequency}, and eyebrow and head gestures have been shown to play a pivotal role in forming negative expressions and wh-questions~\cite{zeshan2004hand,zeshan2004interrogative}.
Furthermore, co-occurring asymmetrical manual signals can convey a high-level of information or other meanings not easily represented by symmetric or sequential manual signals~\cite{sandler2017challenge}.
Hence, excluding these signals leads to translations that are deficient in both semantic and grammatical accuracy.

\begin{figure}[t!]
    \centering
    \includegraphics[width=\linewidth]{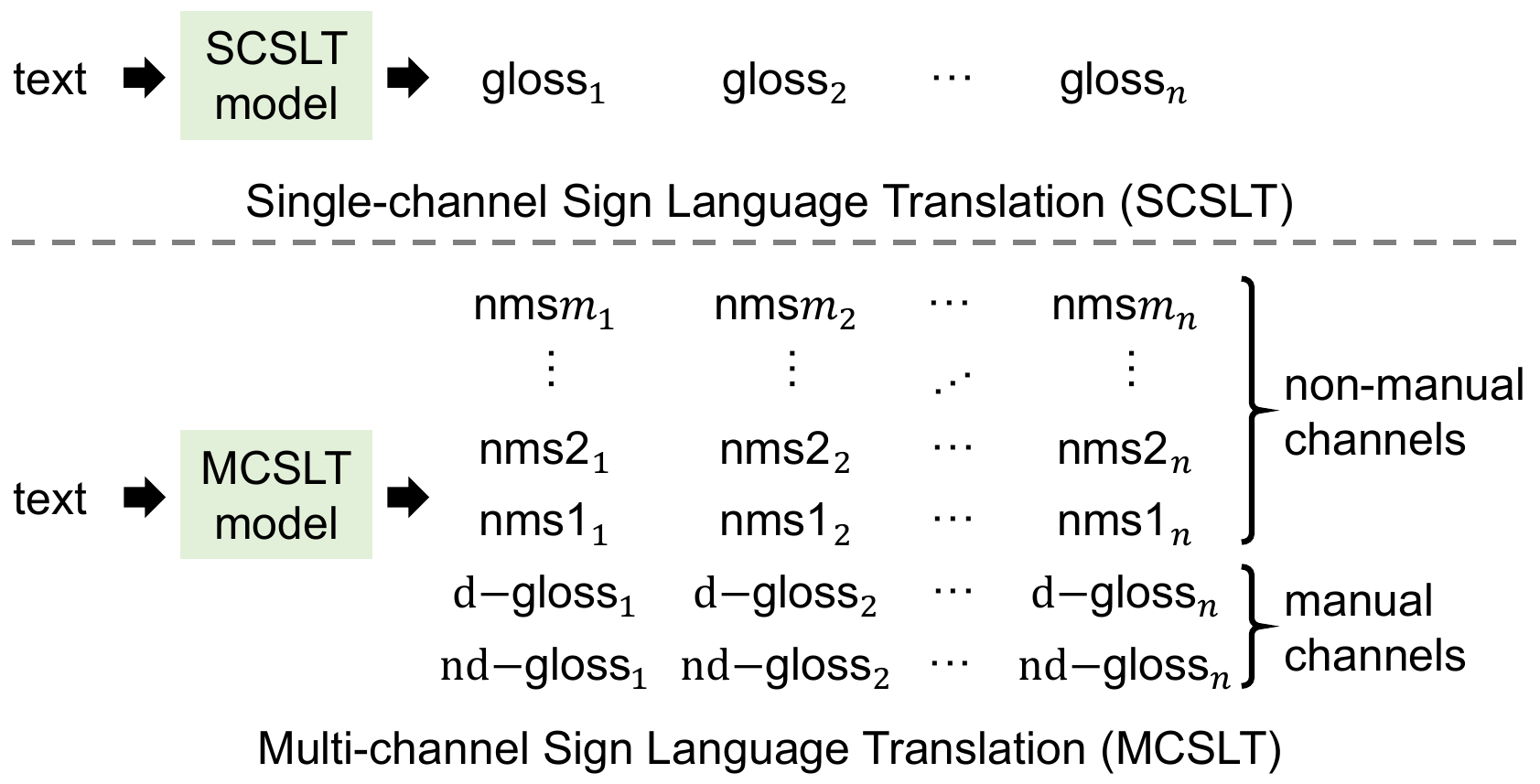}
    \caption{Comparison of SCSLT and MCSLT (``d'': dominant hand, ``nd'': non-dominant hand, and ``nms'': non-manual signal).}
    \label{fig:mcslt}
\end{figure}

To tackle this limitation, we introduce a new task: multi-channel SLT (MCSLT).
First, we redefine the traditional gloss-based SLT that produces only a sequence of manual glosses as single-channel SLT (SCSLT). We then define MCSLT as SLT that predicts signals for \emph{multiple channels}, allowing modeling of concurrent manual and non-manual signals (see Figure~\ref{fig:mcslt} for a visual comparison of the outputs of SCSLT and MCSLT). Note that predicting only two manual channels (for the dominant and non-dominant hands) simultaneously also qualifies as MCSLT. To our knowledge, this study is the first to specifically define and name this approach as MCSLT.
We suspect that the lack of large-scale multi-channel sign language corpora and the absence of a validated metric hindered the emergence of the MCSLT task.

To facilitate meaningful development of MCSLT, we introduce \texttt{SignBLEU}, a new metric designed to capture both sequential and concurrent signals produced by MCSLT.
We tested the proposed metric at the system level by simulating corpus translations and analyzing correlation between text-side \texttt{BLEU} scores and sign-side \texttt{SignBLEU} and other automatic metric scores. \texttt{SignBLEU} showed higher correlation with text-side \texttt{BLEU} scores than other metrics commonly used in SCSLT.
We also showed that at the segment level, \texttt{SignBLEU} has high correlation with human evaluation of translation naturalness and fidelity and of document similarity.
To support future research, we offer initial benchmark MCSLT scores on three sign language corpora.

The key contributions of our paper include:
\begin{itemize}[topsep=4pt]
\setlength\itemsep{0em}
\item The introduction of the multi-channel sign language translation (MCSLT) task, emphasizing the importance of modeling multiple signing channels.
\item The proposal of \texttt{SignBLEU}, a new metric for MCSLT, designed to assess both temporal and concurrent signals.
\item Comprehensive experiments that set baseline MCSLT scores for three sign language corpora and demonstrate that \texttt{SignBLEU} aligns with human evaluation.
\end{itemize}

\section{Related Work}
\label{sec:rel}
We examined the factors that have influenced SLT to date, including corpora, models, and evaluation metrics. We limited analysis to studies translating text to transcribed sign language expressions.

\subsection{Sign Language Corpora}
\begin{figure}[t!]
    \centering
    \includegraphics[width=\linewidth]{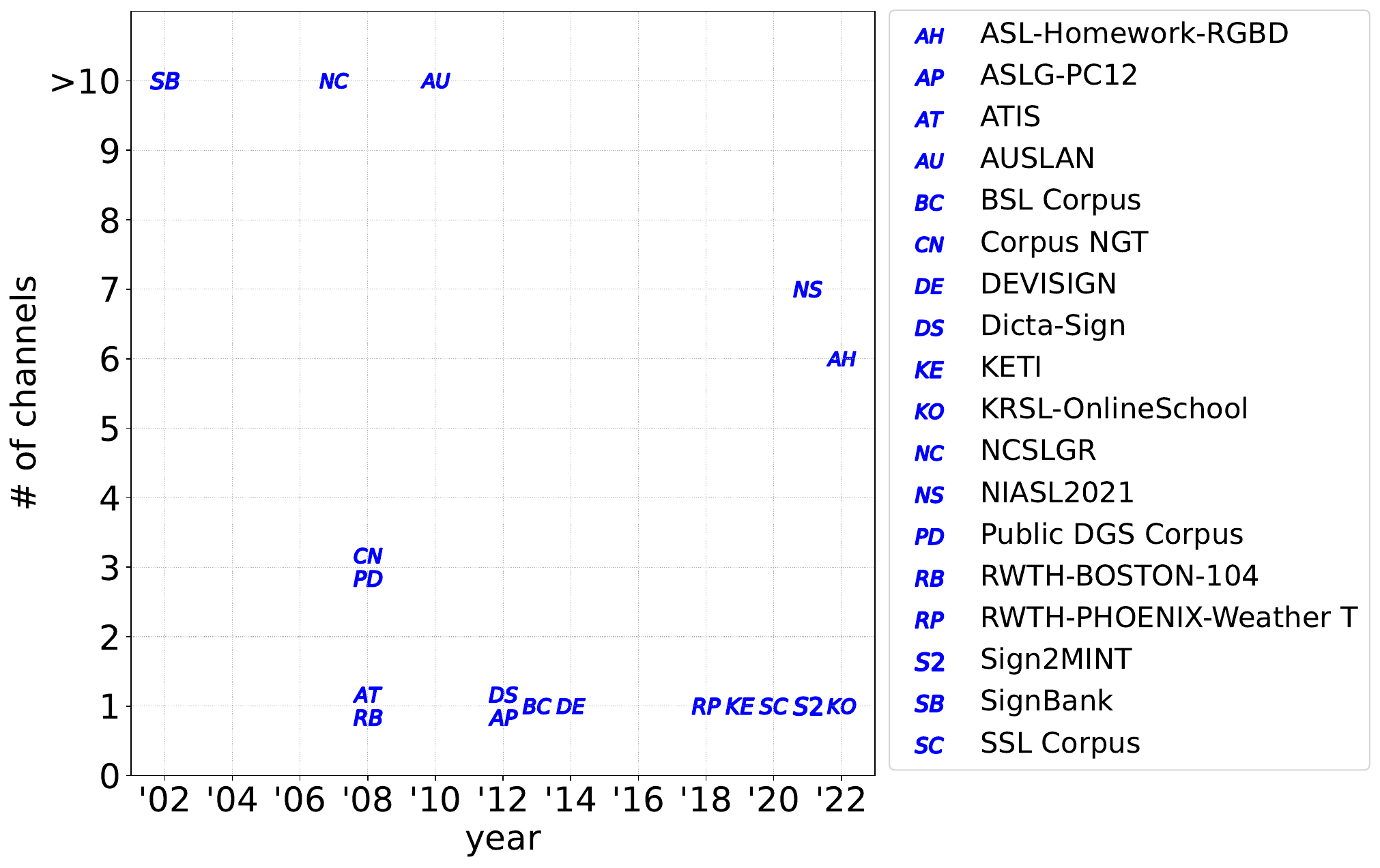}
    \caption{The number of annotated channels of published sign language corpora by year.}
    \label{fig:noc_slt}
\end{figure}

Initially, sign language corpora~\cite{sutton2002signbank,neidle2007signstream,prillwitz08dgs,crasborn:08003:sign-lang:lrec,jbp:/content/journals/10.1075/ijcl.15.1.05joh} were primarily constructed for linguistic analysis of sign language expressions. Therefore, the scale of corpora was relatively small, and there was a tendency to transcribe sign language expressions in as much detail as possible. This detailed transcription was achieved using multi-tier transcription tools like ELAN~\cite{wittenburg-etal-2006-elan} to annotate sign language expressions across multiple signing channels or by developing image-based notations specific to sign language, such as SignWriting~\cite{sutton2000signwriting} and HamNoSys~\cite{hanke2004hamnosys}. However, the introduction of the RWTH-PHOENIX-Weather 2014 T corpus~\cite{camgoz2018neural} signaled the advent of deep-learning-based SLT, prompting a shift towards large-scale data construction. Figure~\ref{fig:noc_slt} illustrates the number of annotated channels in sign language corpora published by year. The scarcity of corpora for MCSLT relative to corpora for SCSLT can be seen clearly from this figure.

\subsection{Sign Language Translation}
\citet{camgoz2018neural} introduced both an NMT-based SLT method and the RWTH-PHOENIX-Weather 2014 T corpus for SLT. SLT performance improved with the adoption of the Transformer~\cite{vaswani2017attention} and with increased use of pre-trained language models as encoders~\cite{camgoz2020sign,miyazaki-etal-2020-machine,de-coster-etal-2021-frozen}. Techniques like data augmentation and multilingual NMT enhanced SLT performance~\cite{moryossef-etal-2021-data,zhu-etal-2023-neural}. However, as mentioned in \cref{sec:intro}, the above methods do not generate non-manual expressions or co-occurring expressions as they continued to be limited to predicting simplistic single-channel signals. To overcome this restriction, \citet{jiang-etal-2023-machine} proposed a text-to-SignWriting method and validated it by categorizing groups within the SignBank corpus~\cite{sutton2002signbank} into being either high-resource or low-resource groups. Yet, with few corpora adopting this transcription methodology, a translation approach applicable to all multi-channel sign language corpora is needed.

\begin{figure*}[t!]
    \centering
    \includegraphics[width=0.925\linewidth]{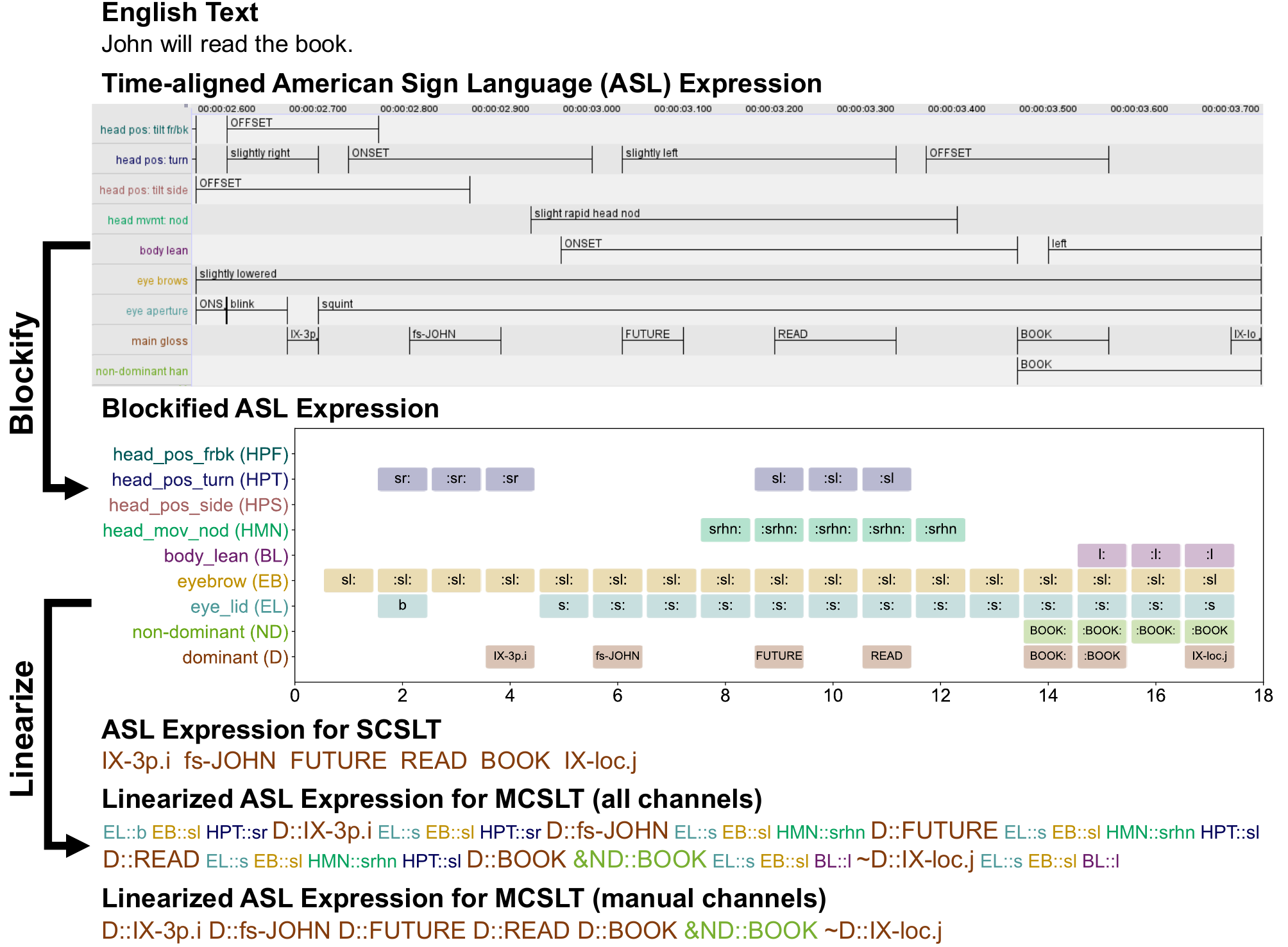}
    \caption{An example of blockification and linearization.}
    \label{fig:blockify_and_linearize}
\end{figure*}

\subsection{Evaluation Metrics for SLT}
\texttt{BLEU}~\cite{papineni02bleu} is the most widely used metric in SLT research. For reproducibility in reporting \texttt{BLEU} scores, many studies have recently turned to using \texttt{sacreBLEU}~\cite{post-2018-call}. \citet{muller23considerations} recommended using \texttt{sacreBLEU} when reporting \texttt{BLEU} scores in SLT and called for reporting metric signatures along with results. As \texttt{BLEU} is precision-based, researchers explored other types of metrics, such as \texttt{ROUGE}~\cite{lin-2004-rouge} and \texttt{METEOR}~\cite{lavie-agarwal-2007-meteor}. Researchers also employed \texttt{chrF}~\cite{popovic-2015-chrf} to measure the character-level $n$-gram \texttt{F}$\mathrm{_1}$ score, and \texttt{TER}~\cite{snover-etal-2006-study} to gauge the edit distance between translated and reference sentences. We argue that any metric for MCSLT needs to be specifically adapted to handle simultaneous signals across multiple channels. Further details on our proposed method are provided in \cref{sec:mcs} and \cref{sec:metric}.

\section{Towards Multi-channel SLT}
\label{sec:mcs}
To model the complex form of multiple signal channels, we introduce two transformations: blockification and linearization.
Blockification converts time-aligned annotation data (e.g., ELAN's EAF format) to a unit-less sequence of co-signed blocks, and linearization converts block data to a simplified text sequence. The block representation should always be used for evaluation, while linearization allows us to apply existing sequence-to-sequence techniques to MCSLT. Figure~\ref{fig:blockify_and_linearize} illustrates these two processes.

\subsection{Blockification}\label{subsec:block}
ELAN and equivalent annotation formats can accurately transcribe sign language expressions down to time alignment and categorical signal attributes, but this representation is too rich to model effectively. Instead, we discretize this representation into a two-dimensional grid of equal-sized gloss blocks.

The ``blockification'' process can be performed in three steps. First, an ordered set of signing channels is identified along with a surjective mapping from annotation tiers to channels (e.g., multiple mouth gesture tiers not containing overlapping annotations may be mapped to a single ``mouth'' channel). Second, the signing timeline is segmented into maximal segments of uninterrupted signing such that barriers between segments correspond to the start or end of at least one signal and no annotation starts or ends within a segment. Third, every non-empty segment is converted to a list of gloss values, sorted according to the given tier-channel mapping and subsequent channel order.

Note that to generate information-rich blocks, we combine annotation tiers by articulator wherever possible when blockifying data. For example, annotations for ``head shake'' and for ``head nod'' can usually be combined into a single ``head'' channel, assuming the gestures never co-occur.

We can formally define the block representation as a gloss-valued $c' \times t'$ matrix:
\begin{align*}
    \mathbf{B}_{c' \times t'}=
    \begin{bmatrix}
        b_{1,1} & \dots & b_{1,t'} \\
        \vdots & \ddots & \vdots \\
        b_{c',1} & \dots & b_{c',t'}
    \end{bmatrix},
\end{align*}
where $c'$ is the number of channels, $t'$ is the number of signal overlap segments, and $b_{ij}$ is either a gloss or $null$. 
The block representation is designed to capture dependent relationships across channels, so that each column is an uninterrupted segment of signing.

As a trivial example, consider the following double-channel data with gloss annotations $g_1$, $g_2$, and $g_3$ on channels $ch_1$ and $ch_2$:

\noindent\rule{\columnwidth}{0.5pt}

$ch_1$: |---------$g_1$---------|

$ch_2$: \hspace{0.4cm}|--$g_2$--| \hspace{0.3cm} |------$g_3$------|

\vspace{-0.2cm}

\noindent\rule{\columnwidth}{0.5pt}

\noindent
The block representation would then be the following $2 \times 5$ matrix:
\begin{align*}
\mathbf{B} &= 
\begin{matrix}
ch_1: & \\
ch_2: &
\end{matrix}
\begin{pmatrix}
g_1: & :g_1: & :g_1: & :g_1 & null \\
null & g_2 & null & g_3: & :g_3
\end{pmatrix}, \\
\end{align*}

\vspace{-0.4cm}

\noindent
where colons denote that a signal is continued to the adjacent column.
See the ``Blockify'' transformation in Figure~\ref{fig:blockify_and_linearize} and \cref{subsec:block_ex} for more information.

Although this process removes duration information, it facilitates the modeling of sign overlap and alignment, which is a more realistic modeling target to progress to from linear gloss sequence prediction.
We consider this representation to be the gold-standard for basic MCSLT and always calculate \texttt{SignBLEU} from block data (see \cref{subsec:tcgrams} for more details).

\subsection{Linearization}\label{subsec:lin}
Inspired by the graph linearization technique of \citet{Bevilacqua_Blloshmi_Navigli_2021}, we transform block sign language expressions into a format that is compatible with existing translation models. Since manual channels typically convey the most meaning, to linearize time-aligned or block data, we first list all manual signals, ordered by signal start time as shown in the ``Linearized ASL Expression for MCSLT (manual channels)'' section of Figure~\ref{fig:blockify_and_linearize}.
We then prefix each signal with a ``D::'', ``ND::'', or ``B::'' to indicate channels for the dominant hand, non-dominant hand, and both hands for two-handed signs, respectively.
To model manual signal overlap, we use the following two prefix tokens:
\begin{itemize}[leftmargin=*]
    \setlength\itemsep{0em}
    \item \textbf{$\sim$}: This token indicates that the current signal starts after but overlaps with the previous manual signal.
    \item \textbf{\&}: This token indicates that the current signal starts at the same time as the previous manual signal (though their endings may differ).
\end{itemize}

Finally, we connect manual signals to co-occurring non-manuals by inserting tokens for non-manual signals directly before or after each manual token with which they overlap, as illustrated in the ``Linearized ASL Expression for MCSLT (all channels)'' portion of Figure~\ref{fig:blockify_and_linearize}. By convention, we interpret a sequence of identical non-manual tokens associated with adjacent manual tokens as one continued non-manual signal, though repetition and continuation can be explicitly modeled by introducing additional special tokens.

Since linearization removes signal start and end alignment, it facilitates easier application of end-to-end MCSLT and allows us to leverage large language models (LLMs) trained on text data from the same cultural region as our target sign language.
Note that we consider the additional information loss in linearization to be an artifact or our current translation technology and not intrinsic to our proposed metric or task.

\section{SignBLEU}  
\label{sec:metric}
In this section, we formally define \texttt{SignBLEU} as a generalization of \texttt{BLEU} to multi-channel block data.

\subsection{Multi-channel N-grams}\label{subsec:tcgrams}
To calculate \texttt{SignBLEU}, we convert block data into $n$-grams using both the column and row dimensions.
We propose using two types of $n$-grams: \emph{temporal grams} for capturing sequential relationships within each signing channel (calculated along rows) and \emph{channel grams} for capturing co-occurring relationships between articulators (calculated within columns).
We denote the order of temporal grams and channel grams by prepending the gram length with ``\texttt{t}'' and ``\texttt{c}'', respectively. E.g., the order of temporal grams of length four is \texttt{t4}.

When calculating temporal grams, continued glosses should be seen as a single element and $null$-valued glosses should be skipped. Channel grams can be calculated as the set of unordered $n$-sized subsets of each column, again skipping $null$ values.

Continuing the double-channel example from \cref{subsec:block}, we can calculate temporal grams of order \texttt{t1} and \texttt{t2} and channel grams of order \texttt{c2} as below.
\begin{center}
    \small
    \begin{tabularx}{0.8\linewidth}{T{0.2} T{0.8}}
        \toprule
        \textbf{Size} & \textbf{Grams} \\
        \midrule
        \texttt{t1} & $\{ch_1g_1\}$, $\{ch_2g_2\}$, $\{ch_2g_3\}$\\
        \texttt{t2} & $\{ch_2g_2, ch_2g_3\}$\\
        \texttt{c2} & $\{ch_1g_1, ch_2g_2\}$, $\{ch_1g_1, ch_2g_3\}$\\
        \bottomrule
    \end{tabularx}
\end{center}

A comprehensive example of the $n$-gram calculation is presented in \cref{subsec:gram_ex}.

Regarding $n$-gram generation from the block and linear representations, it is essential to highlight the following. As mentioned in \cref{subsec:block}, the block representation does not capture duration information---it only represents gloss overlap and alignment. Therefore, raw signing data cannot be reconstructed from the block representation. Similarly, the linear representation cannot fully represent overlap, and conversion from linear to block representation is imperfect. Since we consider the block representation to be the correct representation for MCSLT, reference grams should always be extracted from blockified annotation data, even if hypotheses are generated from linear predictions. If linearized reference data is lifted to a block representation and used to generate $n$-grams, modeling limitations (such as the information loss from linearization) will be ignored and \texttt{SignBLEU} scores will be inflated).

\subsection{Scoring}

After generating temporal and channel grams as above, \texttt{SignBLEU} calculation is analogous to the scoring method for \texttt{BLEU}, with minor adjustments.

First, modified precision is calculated for every $n$-gram type and order, up to the maximum order:
\begin{align*}
    p_n^k = \frac{\sum\limits_{h\in\mathcal{H}}\hspace{0.1cm}\sum\limits_{g\in{gram_n^k(h)}}Count_{clip}(g)}{\sum\limits_{h\raisebox{-.1ex}{\kern-0.05em\ensuremath{\scriptstyle '}}\in\mathcal{H}}\hspace{0.1cm}\sum\limits_{g'\in{gram_n^k(h\raisebox{-.1ex}{\kern-0.05em\ensuremath{\scriptstyle ')}}}}Count(g')},
\end{align*}
where $k\in \{t, c\}$, $p_n^k$ is the precision score of order $n$ and gram type $k$, and $gram_n^k$ is the collection of all $n$-grams of order $n$ and gram type $k$.
Next, a brevity penalty is calculated to penalize short hypotheses.
\begin{align*}
    BP =
    \begin{cases}
    1 & \text{ if } |h|>|r| \\
    e^{(1-|r|/|h|)} & \text{ if } |h| \le |r|
    \end{cases},
\end{align*}
where $|h|$ is the number of annotations in the hypothesis and $|r|$ is the number of annotations in the reference with the most similar length.
Note that we use the raw annotation gloss count (not the block count) to calculate the brevity penalty (e.g, the number of glosses in the toy example from \cref{subsec:block} is three).
Finally, a composite score is created:
\begin{align*}
    \texttt{SignBLEU} = BP \times e^{\bigl(\sum\limits_{n=1}^{n_t} w_n^t\text{log}p_n^t + \sum\limits_{m=2}^{m_c} w_m^c\text{log}p_m^c\bigr)},
\end{align*}
where $w_n^t$ is the weight for the temporal gram precision score of order \texttt{tn}, $w_m^c$ is the weight for the channel gram precision score of order \texttt{cm}, and $n_t$ and $n_c$ are the maximal orders for temporal and channel grams, respectively.

To demonstrate the characteristics of different gram orders, we calculate scores for temporal orders (\texttt{t1}$..$\texttt{t4}) and channel orders (\texttt{c2}$..$\texttt{c4}) for all experiments. Since \texttt{SignBLEU} uses two gram orders, we limit each at four and report up to sixteen order-based variants in our experiments. Similar to \texttt{BLEU}, optimal gram orders and parameter values will depend on the target data and task. Note that we denote \texttt{SignBLEU} with maximal temporal order $n_t$ and channel order $n_c$ as \texttt{SB-t}$n_t$\texttt{c}$n_c$ (e.g., for temporal and channel order $1$, we write \texttt{SB-t1c1}).

\texttt{SignBLEU} can be calculated over all semantically meaningful channels.
However, a manual-only variant of \texttt{SignBLEU} where $n$-grams are extracted only from the manual channels may be appropriate, depending on the target task and data. 
For reproducibility, \texttt{SignBLEU} also provides a signature, similar to \texttt{sacreBLEU}~\cite{post-2018-call} (see~\cref{subsec:metrics}). 
A detailed scoring example is provided in \cref{subsec:scoring_ex}.

\section{Experimental Settings}
\label{sec:exp_set}
This section provides an overview of data, metrics, and implementations used in our experiments.

\subsection{Datasets}
\label{subsec:data}
Due to significant variation across sign language and annotation methodologies, 
it is crucial to assess the proposed \texttt{SignBLEU} on various datasets. To this end, we have selected three sign language corpora. Table~\ref{tab:statistics} contains key statistics of each corpus. Further details on data splits and preprocessing are provided at \url{https://github.com/eq4all-projects/SignBLEU/tree/main/reproducibility}.

\subsubsection{The Public DGS Corpus}
\label{subsubsec:dgs}
The Public DGS Corpus (PDC) is part of the DGS-Korpus project, first introduced by \citet{prillwitz08dgs}. While PDC was not designed as a parallel corpus for training machine translation models, it features comprehensive multi-channel annotations—each hand may be annotated individually, and both mouthings and mouth gestures have annotations—coupled with aligned German and English sentences.  We employ the third release~\citep{dgscorpus3release}, which incorporates the most recent updates as of 2020.

\subsubsection{NIASL2021}
\label{subsubsec:ns21}
\citet{huerta2022kosign} introduced the NIASL2021 corpus (NS21), a large-scale Korean-Korean Sign Language (KSL) parallel corpus for SLT, in 2021. The corpus, based on emergency alerts and weather forecasts, includes non-manual signals from the head, eyebrows, cheeks, and mouth, as well as separate annotations for each manual channel.

\begin{table}[t!]
    \resizebox{\linewidth}{!}{
    \centering
    \begin{tabularx}{1.35\linewidth}{H{0.27} T{0.11} T{0.22} T{0.195} T{0.205}}
        \toprule
         & & \textbf{PDC} & \textbf{NS21} &  \textbf{NCSLGR} \\
        \midrule
        \arrayrulecolor{lightgray} 
        Language Pair & & German-DGS & Korean-KSL & English-ASL \\ \midrule
        \# Instances & \makecell{train\\dev.\\test} & \makecell{61,912\\983\\985} & \makecell{29,980\\1,397\\1,398} & \makecell{1,124\\375\\375} \\ \midrule
        \multicolumn{2}{l}{\makecell[l]{Annotated Channels}} & \makecell{hands,\\mouth} & \makecell{hands,\\head,\\eyebrows,\\cheeks,\\mouth} & \makecell{hands,\\head,\\eyebrows,\\eyes,\\mouth} \\ \midrule
        Vocabulary Size & \makecell{source\\target}& \makecell{19,947\\4,674} & \makecell{4,323\\4,503} & \makecell{1,994\\918} \\ \midrule
        Domain & & \makecell{deaf\\culture} & \makecell{emergency\\alerts} & \makecell{short\\stories} \\
        \arrayrulecolor{black} 
        \bottomrule
    \end{tabularx}
    }
    \caption{Key statistics of sign language corpora.}
    \label{tab:statistics}
\end{table}

\subsubsection{NCSLGR}
\label{subsubsec:bu}
We use the ELAN version of Boston University's The National Center for Sign Language and Gesture Resources corpus (NCSLGR)~\citeplanguageresource{NCSLGR}. We use this corpus as it contains the highest number of annotation tiers, despite its relatively small size.

\subsection{Metrics}
\label{subsec:metrics}
To evaluate the utility of \texttt{SignBLEU}, we pitted it against standard metrics used in SCSLT, including \texttt{BLEU}, \texttt{TER}, \texttt{chrF}, \texttt{METEOR}, and \texttt{ROUGE} (specifically \texttt{ROUGE-L F\textrm{$_1$}}). Non-\texttt{SignBLEU} metrics were calculated on linearized data (see \cref{subsec:lin}).

We calculate two sets of metrics for each experiment. First we calculate scores using all channels and then again for representations of the manual channels only. This allows us to better explore the characteristics of each metric.
We provide the \texttt{SignBLEU} signature\footnote{\relsize{-0.5}off:na||t:3|c:2|dim:1||m:sbleu|sm:exp|eff:n||v:0.1.0}used in our experiments and \texttt{sacreBLEU} version 2.3.1 signatures for \texttt{BLEU}\footnote{\relsize{-0.5}nrefs:1|case:mixed|eff:no|tok:none|smooth:exp}, \texttt{TER}\footnote{\relsize{-0.5}nrefs:1|case:lc|tok:tercom|norm:no|punct:yes|asian:no}, and \texttt{chrF}\footnote{\relsize{-0.5}nrefs:1|case:mixed|eff:yes|nc:6|nw:0|space:no}.

\begin{figure*}[t!]
    \centering
    \includegraphics[width=\linewidth]{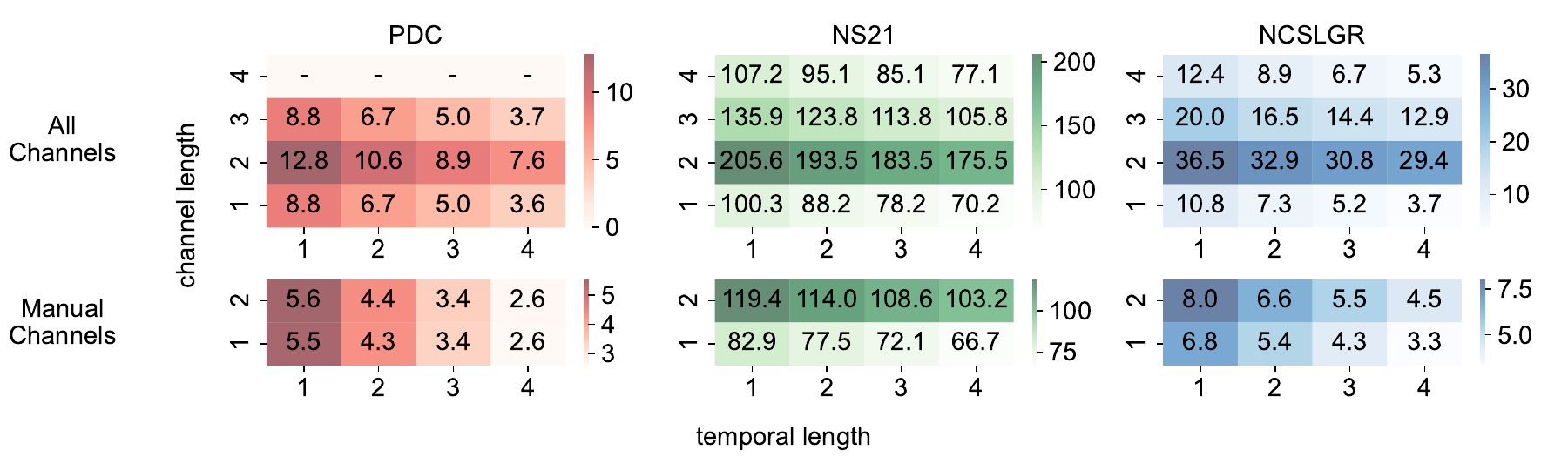}
    \caption{The average $gram_t$ and $gram_c$ counts per sentence by maximum gram order.}
    \label{fig:occurrences}
\end{figure*}

\subsection{Implementation Details}
We fine-tuned pre-trained LLMs on each test corpora. We used \texttt{BLOOM-CLP-German 1.5B} model\footnote{\relsize{-0.5}\url{https://huggingface.co/malteos/bloom-1b5-clp-german}} for German-to-DGS, \texttt{Ko-GPT-Trinity 1.2B} model\footnote{\relsize{-0.5}\url{https://huggingface.co/skt/ko-gpt-trinity-1.2B-v0.5}} for Korean-to-KSL, and \texttt{TinyLlama 1.1B} model\footnote{\relsize{-0.5}\url{https://huggingface.co/TinyLlama/TinyLlama-1.1B-intermediate-step-480k-1T}} for English-to-ASL translation. Each model was fine-tuned and tested on one NVIDIA A100 80GB GPU. We present hyperparameter search results for each model in Table~\ref{tab:hps}.

\section{Experimental Results}
\label{sec:exp_res}

We calculated and report \texttt{SignBLEU} benchmark scores on the test set of each sign language corpus (see~\cref{subsec:e1}). We then validated \texttt{SignBLEU} by analyzing its correlation with text-based \texttt{BLEU} scores at the system level and with human evaluation at the segment level (see~\cref{subsec:e2} and \cref{subsec:e3}). Finally, we developed guidelines on how to interpret \texttt{SignBLEU} scores and offer suggestions for its usage (see~\cref{subsec:e4}).

\begin{table}[t!]
\resizebox{\linewidth}{!}{
\centering
\begin{tabularx}{1.35\linewidth}{H{0.26} T{0.23} *{3}{T{0.17}}}
\toprule
\multirow{2.5}{*}{\textbf{Hyperparameter}} & \multirow{2.5}{*}{\makecell{\textbf{Search}\\\textbf{Space}}} & \multicolumn{3}{c}{\textbf{Pick}}          \\ 
\cmidrule(l){3-5} 
            &                    & \textbf{PDC} & \textbf{NS21} & \textbf{NCSLGR} \\ \midrule
\# Epochs  & $\{1,\ldots,8\}$               & $2$            & $3$             & $6$           \\
LR    & $[10^{-6},10^{-4}]$   & $3.9*10^{-5}$  & $6.0*10^{-5}$   & $8.4*10^{-5}$ \\
Grad. accum.  & $\{4,8,16,32\}$        & $8$            & $8$             & $8$           \\ 
LoRA   &  $\{T,F\}$                    & $F$            & $F$             & $F$           \\
Warm start   &  $\{T,F\}$              & $F$            & $T$             & $F$           \\
Batch size   &  $\{8,16\}$         & $8$            & $16$            & $16$           \\
\bottomrule
\end{tabularx}
}
\caption{Hyperparameter search results.}
\label{tab:hps}
\end{table}

\begin{table}[t!]
\centering
\resizebox{\linewidth}{!}{
\begin{tabularx}{1.45\linewidth}{H{0.165} H{0.145} *{2}{T{0.115}} *{2}{T{0.115}} *{2}{T{0.115}}}
\toprule
\multirow{2.5}{*}{\textbf{Channels}} &\multirow{2.5}{*}{\textbf{Metric}} & \multicolumn{2}{c}{\textbf{PDC}} & \multicolumn{2}{c}{\textbf{NS21}} & \multicolumn{2}{c}{\textbf{NCSLGR}} \\ 
\cmidrule(l{2pt}r{2pt}){3-4} \cmidrule(l{2pt}r{2pt}){5-6} \cmidrule(l{2pt}r{2pt}){7-8}
& & \textbf{Dev.} & \textbf{Test} & \textbf{Dev.} & \textbf{Test} & \textbf{Dev.} & \textbf{Test} \\
\midrule
\arrayrulecolor{lightgray} 
\multirow{17.5}{*}{All} & \texttt{SB-t1c1} & 20.15   & 19.43 & 24.75    & 26.27     & 21.70   & 22.02            \\
                        & \texttt{SB-t1c2} & 14.25   & 13.43 & 21.19    & 22.67     & 18.01   & 18.30            \\
                        & \texttt{SB-t1c3} & 0.00    & 0.00  & 16.02    & 17.60     & 12.27   & 12.45            \\
                        & \texttt{SB-t1c4} & -       & -     & 11.87    & 13.50     & 7.24    & 7.36            \\ \cmidrule{2-8}
                        & \texttt{SB-t2c1} & 9.76    & 9.05  & 13.49    & 15.15     & 10.51   & 10.06            \\
                        & \texttt{SB-t2c2} & 9.87    & 9.13  & 14.90    & 16.50     & 11.82   & 11.55            \\
                        & \texttt{SB-t2c3} & 0.00    & 0.00  & 13.19    & 14.78     & 9.85    & 9.70            \\
                        & \texttt{SB-t2c4} & -       & -     & 10.79    & 12.38     & 6.75    & 6.69            \\ \cmidrule{2-8}
                        & \texttt{SB-t3c1} & 4.89    & 4.20  & 7.52    & 9.17     & 5.88    & 5.06            \\
                        & \texttt{SB-t3c2} & 5.86    & 5.12  & 9.38    & 11.08     & 7.42    & 6.66            \\
                        & \texttt{SB-t3c3} & 0.00    & 0.00  & 9.33    & 10.99     & 7.04    & 6.47            \\
                        & \texttt{SB-t3c4} & -       & -     & 8.36    & 9.96     & 5.43    & 5.08            \\ \cmidrule{2-8}
                        & \texttt{SB-t4c1} & 2.50    & 2.08  & 4.22    & 5.82     & 3.15    & 2.55            \\
                        & \texttt{SB-t4c2} & 3.30    & 2.81  & 5.65    & 7.42     & 4.30    & 3.65            \\
                        & \texttt{SB-t4c3} & 0.00    & 0.00  & 6.13    & 7.87     & 4.51    & 3.94            \\
                        & \texttt{SB-t4c4} & -       & -     & 5.92    & 7.59     & 3.85    & 3.44            \\ 
\arrayrulecolor{black} 
\midrule
\arrayrulecolor{lightgray} 
\multirow{9.5}{*}{Manual} & \texttt{SB-t1c1} & 19.56 & 19.23 & 20.96    & 23.32     & 20.75   & 20.97       \\
                          & \texttt{SB-t1c2} & 0.00  & 0.00  & 20.14    & 22.33     & 5.52    & 6.73       \\
\cmidrule{2-8}
                          & \texttt{SB-t2c1} & 9.54  & 9.21  & 10.47    & 13.07     & 13.71   & 13.23       \\
                          & \texttt{SB-t2c2} & 0.00  & 0.00  & 12.85    & 15.40     & 6.51    & 7.23       \\
\cmidrule{2-8}
                          & \texttt{SB-t3c1} & 4.87  & 4.40  & 5.65    & 8.18     & 9.44    & 8.59       \\
                          & \texttt{SB-t3c2} & 0.00  & 0.00  & 7.69    & 10.40     & 5.93    & 6.08       \\
\cmidrule{2-8}
                          & \texttt{SB-t4c1} & 2.57  & 2.13  & 3.26    & 5.51     & 6.29    & 5.67       \\
                          & \texttt{SB-t4c2} & 0.00  & 0.00  & 4.66    & 7.23     & 4.70    & 4.68       \\
\arrayrulecolor{black} 
\bottomrule
\end{tabularx}
}
\caption{MCSLT Benchmark scores.}
\label{tab:benchmarks}
\end{table}

\subsection{MCSLT Benchmark Scores}
\label{subsec:e1}
We present MCSLT benchmark scores for the test sets in Table~\ref{tab:benchmarks}. As mentioned in \cref{sec:intro}, these are \emph{initial} MCSLT benchmarks, and we share them with the hope that we can encourage further research and advancements in MCSLT. 

We report all-channel and manual-channel \texttt{SignBLEU} scores up to gram order \texttt{t4c4}, resulting in 24 different metrics. Naturally, it can be challenging to determine which metrics to prioritize for each corpus.
We suggest analyzing both gram frequencies and annotation methodologies as a good starting point. Figure~\ref{fig:occurrences} presents the average $n_t$-gram and $n_c$-gram counts per sentence categorized by temporal and channel lengths. These counts not only reveal the characteristics of each corpus, but can also provide a glimpse into which temporal and channel levels it would be beneficial to focus on for each corpus.

\begin{table}[t!]
\centering
\resizebox{1\linewidth}{!}{
\begin{tabularx}{1.45\linewidth}{H{0.164}  H{0.001} H{0.151} *{2}{T{0.114}} *{2}{T{0.114}} *{2}{T{0.114}}}
\toprule
\multirow{2.5}{*}{\textbf{Channels}} & \multicolumn{2}{l}{\multirow{2.5}{*}{\textbf{Metric}}} & \multicolumn{2}{c}{\textbf{PDC}} & \multicolumn{2}{c}{\textbf{NS21}} & \multicolumn{2}{c}{\textbf{NCSLGR}} \\ 
\cmidrule(l{2pt}r{2pt}){4-5} \cmidrule(l{2pt}r{2pt}){6-7} \cmidrule(l{2pt}r{2pt}){8-9}
& & & $\boldsymbol\rho$ & $\boldsymbol\tau$ & $\boldsymbol\rho$ & $\boldsymbol\tau$ & $\boldsymbol\rho$ & $\boldsymbol\tau$ \\ \midrule
\arrayrulecolor{lightgray} 
\multirow{28}{*}{All}& \multicolumn{3}{l}{Existing Metrics}      &     &     &      &     &         \\ 
 & & \texttt{BLEU-1} & .206 & .140 & .153 & .104 & .128 & .083 \\ 
 & & \texttt{BLEU-2} & .258 & .178 & .228 & .153 & .276 & .184 \\ 
 & & \texttt{BLEU-3} & .264 & .182 & .272 & .183 & .450 & .302 \\ 
 & & \texttt{BLEU-4} & .267 & .184 & .251 & .171 & .457 & .305 \\ 
 & & \texttt{chrF} & .127 & .085 & .127 & .085 & .205 & .136 \\ 
 & & \texttt{METEOR} & .229 & .154 & .199 & .135 & .277 & .180 \\ 
 & & \texttt{ROUGE} & .225 & .151 & .196 & .133 & .242 & .160 \\ 
 & & \texttt{1-TER} & .081 & .055 & .099 & .065 & -0.021 & -0.013 \\ 
\cmidrule{2-9}
& \multicolumn{2}{l}{\texttt{SignBLEU}} &      &     &     &      &     &         \\
 & & \texttt{SB-t1c1} & .207 & .140 & .186 & .125 & .248 & .163 \\ 
 & & \texttt{SB-t1c2} & .238 & .169 & .211 & .142 & .260 & .174 \\ 
 & & \texttt{SB-t1c3} & .054 & .044 & .193 & .130 & .325 & .226 \\ 
 & & \texttt{SB-t1c4} & - & - & .156 & .106 & .400 & .321 \\ \cmidrule{2-9}
 & & \texttt{SB-t2c1} & .229 & .173 & .205 & .138 & .389 & .261 \\ 
 & & \texttt{SB-t2c2} & .259 & .201 & .230 & .154 & .372 & .248 \\ 
 & & \texttt{SB-t2c3} & .064 & .053 & .217 & .146 & .359 & .251 \\ 
 & & \texttt{SB-t2c4} & - & - & .178 & .120 & .403 & .325 \\ \cmidrule{2-9}
 & & \texttt{SB-t3c1} & \underline{.401} & \underline{.326} & .189 & .127 & .543 & .416 \\ 
 & & \texttt{SB-t3c2} & \underline{.410} & \underline{.334} & .211 & .142 & .543 & .414 \\ 
 & & \texttt{SB-t3c3} & .064 & .053 & .220 & .149 & .537 & .417 \\ 
 & & \texttt{SB-t3c4} & - & - & .197 & .134 & .451 & .367 \\ \cmidrule{2-9}
 & & \texttt{SB-t4c1} & \underline{.390} & \underline{.318} & .172 & .117 & \underline{.648} & \underline{.525} \\ 
 & & \texttt{SB-t4c2} & \underline{.389} & .318 & .185 & .126 & \underline{.648} & \underline{.524} \\ 
 & & \texttt{SB-t4c3} & .064 & .053 & .201 & .137 & .618 & \underline{.498} \\ 
 & & \texttt{SB-t4c4} & - & - & .207 & .142 & .466 & .380 \\ 
\arrayrulecolor{black} 
\midrule
\arrayrulecolor{lightgray} 
\multirow{20}{*}{Manual} & \multicolumn{3}{l}{Existing Metrics}      &     &     &      &     &         \\ 
 & & \texttt{BLEU-1} & .186 & .125 & .221 & .148 & .362 & .249 \\ 
 & & \texttt{BLEU-2} & .264 & .181 & \underline{.309} & \underline{.211} & .604 & .431 \\ 
 & & \texttt{BLEU-3} & .274 & .189 & .305 & .209 & .614 & .441 \\ 
 & & \texttt{BLEU-4} & .279 & .193 & .301 & .207 & .613 & .439 \\ 
 & & \texttt{chrF} & .146 & .099 & .255 & .173 & .445 & .308 \\ 
 & & \texttt{METEOR} & .224 & .151 & .262 & .178 & .488 & .335 \\ 
 & & \texttt{ROUGE} & .208 & .140 & .236 & .161 & .372 & .251 \\ 
 & & \texttt{1-TER} & .081 & .054 & .122 & .081 & -0.026 & -0.017 \\ 
\cmidrule{2-9}
& \multicolumn{2}{l}{\texttt{SignBLEU}} &      &     &     &      &     &         \\
 & & \texttt{SB-t1c1} & .191 & .128 & .238 & .161 & .369 & .252 \\ 
 & & \texttt{SB-t1c2} & .062 & .050 & .235 & .160 & .368 & .292 \\ \cmidrule{2-9}
 & & \texttt{SB-t2c1} & .320 & .255 & \hl{\textbf{.326}} & \hl{\textbf{.223}} & \underline{.623} & .477 \\ 
 & & \texttt{SB-t2c2} & .084 & .069 & \underline{.320} & \underline{.218} & .449 & .364 \\ \cmidrule{2-9}
 & & \texttt{SB-t3c1} & \hl{\textbf{.422}} & \hl{\textbf{.345}} & \underline{.319} & \underline{.219} & \hl{\textbf{.703}} & \hl{\textbf{.570}} \\ 
 & & \texttt{SB-t3c2} & .084 & .069 & \underline{.319} & \underline{.218} & .475 & .387 \\ \cmidrule{2-9}
 & & \texttt{SB-t4c1} & .389 & \underline{.318} & .280 & .204 & \underline{.689} & \underline{.561} \\ 
 & & \texttt{SB-t4c2} & .070 & .057 & .280 & .204 & .485 & .395 \\ 
\arrayrulecolor{black} 
\bottomrule
\end{tabularx}
}
\caption{System level correlations of text-side \texttt{BLEU} with multiple sign language metrics. We highlight the \hl{\textbf{top-1}} and \underline{top-5} highest correlation scores for readability.}
\label{tab:corr_textbleu}
\end{table}

\subsection{Correlation with Text-side BLEU}
\label{subsec:e2}
Inspired by the success of backtranslation for translation evaluation and given that our three test sets have both text-side and sign-side annotations, we measured the correlation between corpus \texttt{BLEU} scores on the text side and various corpus metric scores on the sign side.
To do so, we make two assumptions: supplied text translations of sign language data are of high quality and corpus \texttt{BLEU} is reliable for system evaluation.

We simulated translation systems using randomly sampled hypothesis and reference sentences. For each corpus and simulation run, we sampled 200 instances, splitting the samples into two sets of 100 instances. We used one set as reference translations and one set as hypothesis translations. We then scored the simulated system with each metric. We repeated these sampling and scoring steps 10,000 times to get 10,000 system scores. Finally, we calculated rank correlation (using Spearman's Rho and Kendall's Tau-b) between each sign-side metric and the text-side \texttt{BLEU} scores. Unlike most system-level analysis, this simulation does not compare system performance over the same instances. However, due to the law of large numbers, this will not matter given enough samples.

\begin{figure*}[t!]
    \centering
    \begin{subfigure}{0.495\textwidth}
        \centering
        \includegraphics[width=\linewidth]{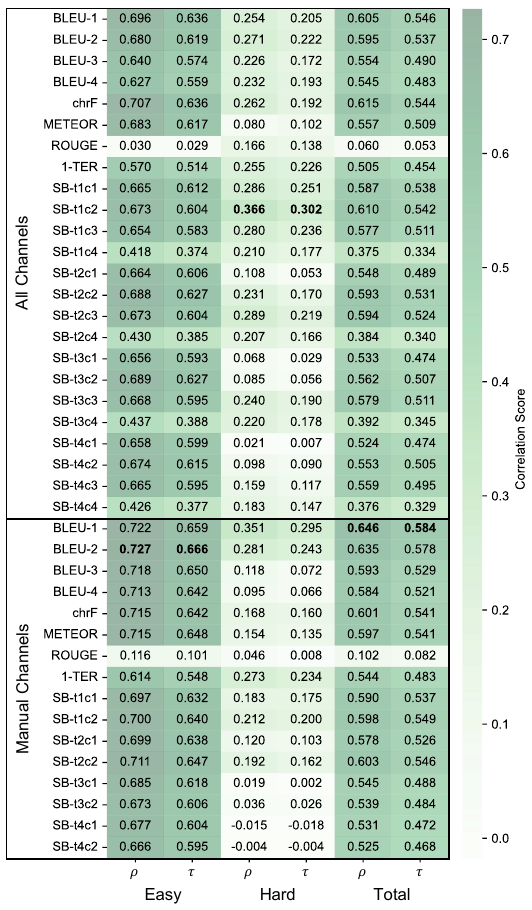}
        \caption{similarity ranking}
        \label{fig:simrank_results}
    \end{subfigure}
    \hfill
    \begin{subfigure}{0.495\textwidth}
        \centering
        \includegraphics[width=\linewidth]{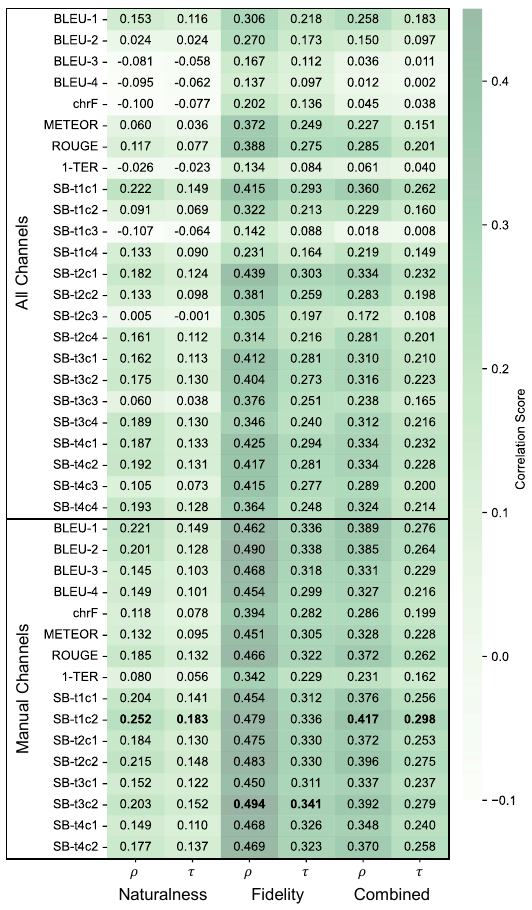}
        \caption{direct assessments}
        \label{fig:natfid_results}
    \end{subfigure}
    \caption{Correlations with human judgments. Highest scores are highlighted in bold for readability.}
    \label{fig:human_judgments}
\end{figure*}

Table~\ref{tab:corr_textbleu} provides a summary of the correlation results. All three datasets showed higher correlation with a \texttt{SignBLEU} variant than with other metrics. We can also see features of each dataset reflected in the different gram order scores.
For example, the difference between single-channel and double-channel manual \texttt{SignBLEU} correlation for PDC is likely due to how most PDC manual signs are annotated for only a single channel. Contrast this with NS21 which contains mostly two-handed (symmetric and asymmetric) manuals and shows little difference between single-channel and double-channel manual \texttt{SignBLEU} scores.

\subsection{Correlation with Human Judgment}
\label{subsec:e3}
We performed two additional segment-level experiments to explore agreement with human evaluation.

\subsubsection{Task \#1: Ranking}
\label{subsubsec:simrank}

We examined the correlation between sentence-level similarity rankings by \texttt{SignBLEU} and human-based similarity rankings on NS21, using a set of 147 questions.
For each question, a reference sign language video and four candidate videos were provided. Most videos were between ten and twenty seconds long. Evaluators ranked the candidate videos by similarity to the reference video and were instructed to base their rankings on meaning similarity first and signing flow and vocabulary use second. Out of the 147 questions, 115 were easier to rank (termed ``\textbf{Easy}'') as the candidate videos were randomly sampled from both the same and different domains as the reference video, leading to larger semantic differences between candidate videos. The remaining 32 questions, labeled ``\textbf{Hard}'', posed a greater challenge. Their candidate videos were intentionally sampled to have a unigram gloss precision of over 90\%, requiring close scrutiny to determine similarity rankings.
While this is not a translation task, it provides insight into the MCSLT and \texttt{SignBLEU}.

Four native signers individually ranked candidate videos. We then aggregated their responses into labels, allowing for ties. Metric-based similarity was calculated with our test metrics applied to existing annotations for each video. To reduce bias, candidate videos were selected with one additional criterion--all five videos had to have either the same signer wearing the same outfit or have different signers.

Figure~\ref{fig:simrank_results} presents the rank correlation as a heatmap. Correlation was computed using both Spearman's Rho ($\rho$) and Kendall's Tau-b ($\tau$). Manual \texttt{BLEU-1} and manual \texttt{BLEU-2} showed the highest agreement with human scores on the total ranked dataset and on the ``Easy'' subset, respectively. Post-evaluation interviews with the four evaluators revealed that for the ``Easy'' ranking task, non-manuals and co-occurring signs could be completely ignored and almost all candidate videos could still be correctly ranked. Thus, it makes sense that a simple \texttt{BLEU-1} or \texttt{BLEU-2} score would perform well for this task.
On the other hand, rankings from the ``Hard'' subset showed higher correlation with all-channel \texttt{SignBLEU} scores, especially scores of channel order \texttt{c2}, \texttt{c3}, and \texttt{c4}. This suggests that evaluation of multiple channels, including non-manual channels, was required to effectively rank the candidates.

It is worth noting that since NS21 was constructed from manually-translated emergency alerts and weather broadcasts, non-manual signals may play a secondary role to manual signals, in contrast to collected from spontaneous signing.

\subsubsection{Task \#2: Direct Assessments}
\label{subsubsec:natfid}

Since our primary objective was to validate \texttt{SignBLEU} as a metric for machine translation, we collected native signer direct assessments of automatic translation results and compared them with metric scores. We randomly selected 53 instances from the development subset of NS21 and generated translations for each instance using the same model used to report NS21 benchmark scores. To avoid bias introduced by the influence of an avatar representation, we hired two experienced signers to create signing videos based on the translations by re-signing the predicted multi-channel glosses. Eighteen evaluators then scored each video. For each instance, each evaluator first watched the re-signed video and scored it for naturalness. They then viewed one of the correct reference translation videos and scored the re-signed video for fidelity. All training and evaluation was conducted in sign language.
Naturalness and fidelity were both evaluated on eleven-point Likert scales labeled uniformly from 0 to 100.

Again, we analyzed correlation between evaluator and metric-based scores using Spearman's Rho ($\rho$) and Kendall's Tau-b ($\tau$). Results are displayed in Figure~\ref{fig:natfid_results}. Note that we used z-scores calculated separately over each evaluator's naturalness and fidelity scores for correlation analysis. ``\textbf{Combined}'' was calculated from the mean of naturalness and fidelity z-scores.

All metrics showed higher correlation with fidelity than with naturalness.
This aligns with results from the ``Easy'' subset of the similarity ranking experiment, where metrics evaluated on manual channels demonstrated higher correlations than those on all channels.
These experimental results further illustrated that NS21 is more biased towards manual information.
Overall, we found that \texttt{SignBLEU} outperformed existing metrics.
Interestingly, manual \texttt{SB-t1c2}, which emphasizes co-occurring signs, showed the highest correlation with human-scored naturalness, and manual \texttt{SB-t3c2}, which captures a sequential relationship in addition to co-occurring relationships, showed the best correlation with human-scored fidelity.

\subsection{SignBLEU Guideline}
\label{subsec:e4}
To use \texttt{SignBLEU}, an appropriate max gram order must be selected. One can simply use the \texttt{t1c2} variant due to the high number of temporal and channel grams of this order, as seen in Figure~\ref{fig:occurrences}. This variant also showed high correlation with human judgment on NS21. However, it showed poor and mediocre correlation with manual-only and all-channel PDC, respectively.

If human evaluation is available for one's corpus, it should be utilized to find appropriate gram orders.
If it is not available, but text-side translations for your data are, we recommend performing correlation analysis with text-side \texttt{BLEU}, as in \cref{subsec:e2}.

To help with gram order and other parameter selection, we will publish additional analysis online at \url{https://github.com/eq4all-projects/SignBLEU}.

\section{Conclusions and Future Work}
\label{sec:con}
In this study, we proposed a new gloss-based sign language translation (SLT) task that we termed multi-channel sign language translation (MCSLT).
MCSLT refers to any SLT that generates gloss predictions across multiple signal channels. 
We then proposed and validated a new metric, \texttt{SignBLEU}, for MCSLT evaluation.
We hope that more SLT research will adopt the multi-channel approach, and we will continue to evaluate and refine \texttt{SignBLEU} as an open-source solution to MCSLT evaluation.

\section{Acknowledgment}
This work was partly supported by the Institute for Information and communications Technology Promotion (IITP) grant funded by the Ministry of Science and ICT (MSIT, Korea) (No. 2022-0-00010, Development of Korean sign language translation service technology for the deaf in medical environment) and the Technology Innovation Program funded by the Ministry of Trade, Industry \& Energy (MOTIE, Korea) (No. 20014406, Development of interactive sign language interpretation service based on artificial intelligence for the hearing impaired).

\section{Ethical Considerations}
\label{sec:eth}
The study's protocol was approved by the Korea National Institute for Bioethics Policy (IRB No. P01-202310-01-014). All participants were informed about the nature, purpose, procedures, potential risks, and benefits of the research. They gave their voluntary consent to participate, ensuring they felt no pressure. Moreover, they were made aware of their right to withdraw from the study at any time without any repercussions.

To ensure participant compensation, we collected certain personal information. However, upon completing the compensation-related administrative processes, all personal data was destroyed. For the sake of data security, access to the evaluation data was restricted to the authors, all of whom are registered researchers under the research plan sanctioned by the IRB.

Our study, aimed at comparing and assessing sign language videos, necessitated the inclusion of deaf individuals who use sign language as their primary language of communication. Every step, ranging from recruitment and guideline explanation to the evaluation itself, was communicated in sign language to ensure clear communication. Risks for participants were kept to a minimum.

Participants spent a maximum of 2 hours in the study, spanning the time from introduction to the evaluation guidelines through to the completion of the actual evaluation. As compensation for their time and insights, they received payment exceeding the national minimum wage.

\section{Limitations}
\label{sec:lim}

\begin{itemize}[leftmargin=*]
    \item This study was focused on developing and validating a metric for automatic evaluation of MCSLT. Although the translation model used in our experiments was optimized through hyperparameter search, the reported scores should be considered only as preliminary benchmark scores for MCSLT, and we do not consider our modeling approach itself to be a technical contribution.
    \item Linearized sign language expressions and equivalent multi-channel sign language expressions predicted by MCSLT models are human-readable but are not directly viewable as sign language expressions. Therefore, to conduct human evaluations of naturalness and fidelity, we presented the output as a sign language video re-signed by native signers so that the evaluator would not be negatively biased by either raw visualization or by an avatar representation. However, this approach required re-signing the predicted MCSLT \emph{exactly}, which proved extremely difficult. While we cannot guarantee that we were able to eliminate all production bias, we conducted several rounds of review for each video to remove extraneous signals. Since most errors could be identified quickly, the bigger challenge was simply the energy- and time-cost of re-signing. Since the synthetic utterances included many small ``errors'', signers had to practice each utterance before filming and most videos were re-filmed at least once.
    Due to these costs, we advise against using this approach and emphasize the need to find a better solution to isolated human evaluation of SCSLT and MCSLT results, unbiased by avatar and other production methodologies.
    \item Though we provided some interpretation as to why certain max gram order variants performed well or poorly, it is important to recognize that the optimal choice of max gram order will depend on the target corpus and the user's specific objectives.
    \item All corpora used in this study contain different language pairs (German-DGS, Korean-KSL, and English-ASL). Due to this, there were ethical and accessibility-related limitations to performing user evaluations for every corpus. To alleviate this to some extent, and inspired by the practice of assessing quality through backtranslation in sign language production research, we calculated correlation with text side \texttt{BLEU} score, attempting to provide as objective a validation as possible for all sign language corpora.
\end{itemize}

\nocite{*}
\section{Bibliographical References}\label{reference}

\bibliographystyle{lrec_natbib}
\bibliography{lrec-coling2024}

\section{Language Resource References}
\label{lr:ref}
\bibliographystylelanguageresource{lrec_natbib}
\bibliographylanguageresource{languageresource}

\appendix
\renewcommand\thesection{Appendix \Alph{section}} 
\renewcommand\thesubsection{\Alph{section}.\arabic{subsection}}

\section{Calculation Example}
This section illustrates how \texttt{SignBLEU} is calculated using two example documents, document 1 and document 2, as seen in figure~\ref{fig:elan_ex} (top and bottom, respectively).
This appendix is provided to supplement the explanations for blockification from \cref{sec:mcs} and both \texttt{SignBLEU} scoring (including gram creation) from \cref{sec:metric}.

See \cref{subsec:block_ex} the example blockification calculation and \cref{subsec:gram_ex} for gram calculation.
Scoring is an extension of the modified $n$-gram precision scoring from the original \texttt{BLEU} algorithm, and the calculation for this example is covered briefly in \cref{subsec:scoring_ex}.

The two examples shown here are synthetic documents containing some degree of gloss overlap.
Manual tiers are ``both'' for signals using both hands and ``right'' for right-hand only signals.
All other tiers are non-manual tiers.

\subsection{Multi-Channel Blocks}
\label{subsec:block_ex}
Blocks can be generated iteratively using annotation start and end times.

Let $G$ denote a collection of gloss annotations; let $T=\{t_i\}$ denote the collection of all gloss start and end times, de-duplicated and sorted in ascending order; and let $g.start$, $g.end$, $g.tier$, and $g.name$ denote annotation start, end, tier, and gloss name for annotation $g$.
Also assume that we have a mapping $M: tier \mapsto channel$ that maps tiers to target channels.
$M$ need not be injective as we may want to map multiple tiers to the same signal channel.
The block representation $B$ of a document can then be calculated using algorithm~\ref{alg:block}.

\begin{algorithm}
\caption{$\text{blockify}(G, T, M): B$}
\label{alg:block}
\begin{algorithmic}[1]
\State $n \gets (|T|-1)$
\State $B \gets \{\}$
\For{$i \in \{1...n\}$}
    \State $\triangleright$ Initialize block dictionary
    \State $block \gets \{\}$
    \State $gs \gets \{g | g \in G, $
    \Statex \hspace{1.5cm} $g.start \le t_i < t_{i+1} \le g.end \}$
    \For{$g \in gs$}
        \State $\triangleright$ Denote continuation
        \State $prefix \gets \text{``}:\text{''}\textbf{ if } g.start < t_i \textbf{ else } \text{`` ''}$
        \State $suffix \gets \text{``}:\text{''} \textbf{ if } g.end > t_{i+1} \textbf{ else } \text{`` ''}$
        \State $name \gets prefix + g.name + suffix$
        \State $channel \gets M(g.tier)$
        \State $block[channel] \gets name$
    \EndFor
    \If{$|block| > 0$}
        \State $B.append(block)$
    \EndIf
\EndFor
\State \textbf{Return} $B$
\end{algorithmic}
\end{algorithm}

\noindent
This generates a sequence of blocks, where each block maps channels to gloss names.
By convention, we add a key-value pair for each missing channel, mapping the channel to \texttt{null}.
Glosses may be renamed by pre- or post-pending a special symbol (shown here as $\text{``}:\text{''}$) to the gloss name to mark continuation from the previous or to the next block, respectively.
Continuation identifiers are used to calculate intra-channel (temporal) grams directly from the block representation and are used in several \texttt{SignBLEU} variants that we are still developing and plan to release in the future.
Given any fixed channel order $\gamma$, we can represent a block sequence as a block matrix by converting each block to a column vector with values ordered by the order of their keys in $\gamma$.
We consider the block matrix synonymous with the block sequence representation and refer to them both as block representations.

See Table~\ref{tab:block_ex} for example block representations of both ELAN examples from Figure~\ref{fig:elan_ex}.

\subsection{Temporal and Channel Grams}
\label{subsec:gram_ex}
Given annotation data represented as a block matrix, $n$-grams can be easily calculated by extracting $n$ adjacent glosses from each row across blocks (temporal grams) and sets of size $n$ of non-\texttt{null} glosses across channels from within each column (channel grams).

\subsubsection{Temporal Grams}
Given a block matrix $B$, temporal grams can be calculated as

\begin{align*}
    &\bigcup\limits_{row\in B} gram_n(\{b \ | \ b \in row, b \ne \texttt{null}, \neg pre(b)\}),
\end{align*}

where $gram_n$ is the standard $n$-gram function and $pre$ is true if and only if there is a continuation prefix.
All experiments from this study used this simple implementation to extract temporal grams of order \texttt{t1}..\texttt{t4} from each channel.
Since channels may be constructed from multiple tiers during blockification, extracting temporal grams from the block representation may be easier than from the original time-aligned annotation representation.
Simply collect adjacent non-\texttt{null} glosses, skipping those that start with continuation markers.

We are experimenting with including whitespace (\texttt{null} values) and with weighting based on the number of blocks a single signal spans, and we may introduce parameters to allow for different temporal gram calculations in the future.

\subsubsection{Channel Grams}
Channel grams are intra-block, inter-channel grams (i.e., constructed from within a single block column).
However, since channels have no inherent order, channel grams of size $n$ from a given block are the set of all $n$-length subsets of non-\texttt{null}-annotations from that block.
When calculating both temporal and channel grams, we skip channel grams of order \texttt{c1} since the high level of overlap between temporal grams of order \texttt{t1} and channel grams of order \texttt{c1} led to worse performance.

\subsubsection{2D Grams}
We experimented with two-dimensional grams constructed from both the temporal and channel dimensions, but the combination of separate temporal and channel grams performed better than the implementations of two-dimensional grams that we tested.
Two-dimensional grams also suffer from two other challenges: they are more sensitive to small alignment changes and they lead to a much higher computational complexity due to the increased number of unique grams.
We plan to continue improving the two-dimensional implementation as a possible future improvement.

\subsection{Scoring}
\label{subsec:scoring_ex}
As stated above, scoring is analogous to that of the original \texttt{BLEU} algorithm, with adjustments to handle multiple types of $n$-grams.

We found that weighting each gram order and type evenly performed well in our original experiments and recommend doing so as a safe starting point.
We calculate the brevity penalty using the number of annotations included in the calculation, which can be calculated from the block representation by counting glosses that do not start with a continuation prefix.
We tested several other variants, including the number of blocks, the number of glosses (with or without a continuation prefix), and the number of blocks containing manual glosses, but the simple annotation count performed the best in our initial experiments.

For the hypothesis and reference presented in Figure~\ref{fig:elan_ex}, the modified precision for orders \texttt{t1}, \texttt{t2}, \texttt{t3}, and \texttt{c2} are as follows:
\begin{center}
    \begin{tabularx}{0.8\linewidth}{T{0.3} T{0.7}}
        \toprule
        \textbf{Order} & \textbf{Score} \\
        \midrule
        \texttt{t1} & 0.368421 \\
        \texttt{t2} & 0.266667 \\
        \texttt{t3} & 0.181818 \\
        \texttt{c2} & 0.625 \\
        \bottomrule
    \end{tabularx}
\end{center}

Finally, we can calculate the raw aggregate score, the brevity penalty, and the final \texttt{SignBLEU} score:
\begin{center}
    \begin{tabularx}{0.8\linewidth}{T{0.3} T{0.7}}
        \toprule
        & \textbf{Score} \\
        \midrule
        Raw & 0.325056 \\
        BP & 0.768621 \\
        \texttt{SignBLEU} & 0.249844 \\
        \bottomrule
    \end{tabularx}
\end{center}

\onecolumn
\begin{landscape}

\begin{figure}[p]
    \centering
    \includegraphics[width=\linewidth]{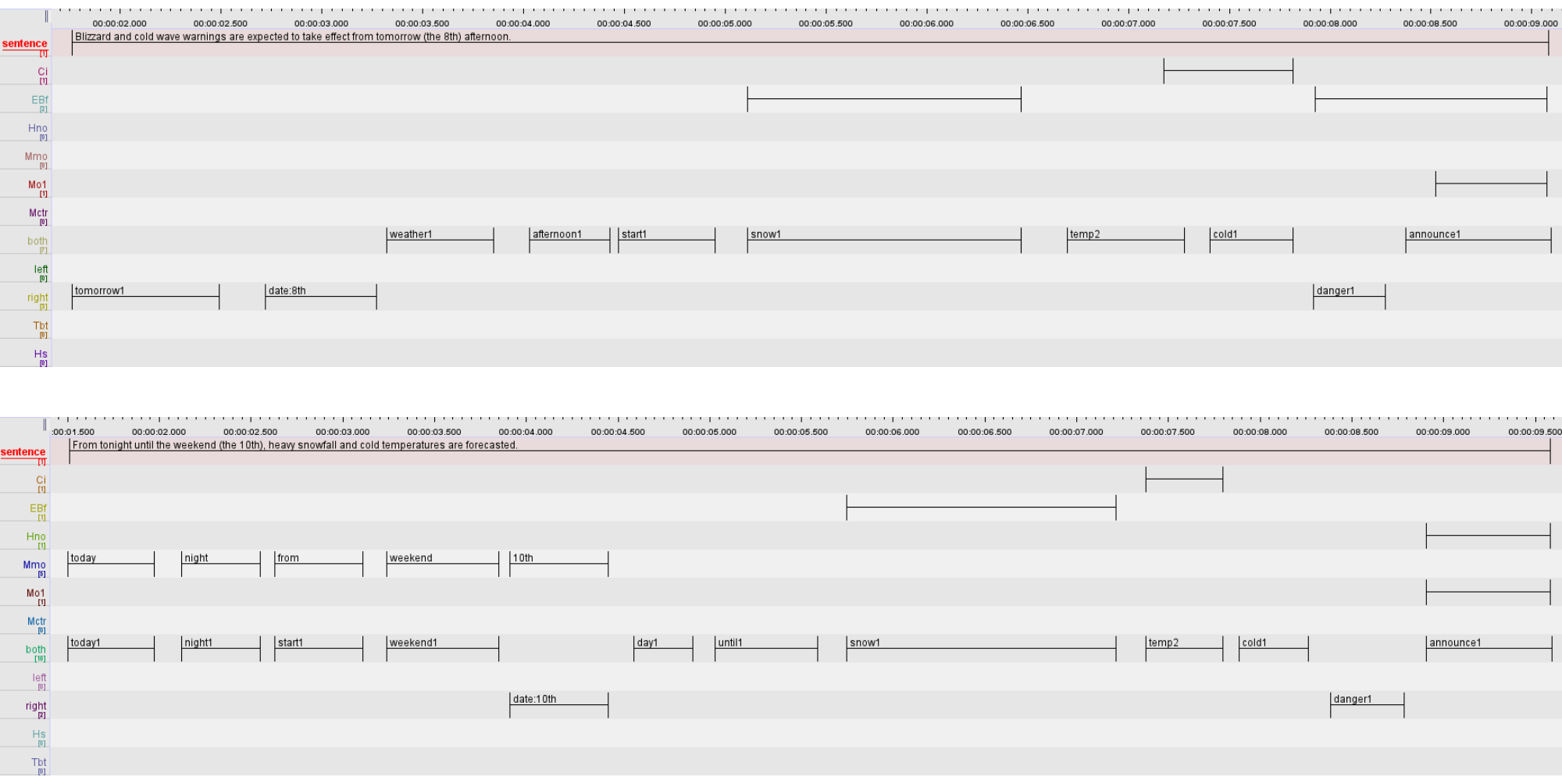}
    \caption{Sample ELAN instances.}
    \label{fig:elan_ex}
\end{figure}

\newpage

\renewcommand{\arraystretch}{1.6}

\begin{table}[p]
\centering
\small
\resizebox{\columnwidth}{!}{
\begin{tabular}{ll *{22}{p{1.6cm}}}
    \toprule
    \multirow{2}{*}{\textbf{Doc}} & \multirow{2}{*}{\textbf{Channel}} & \multicolumn{22}{c}{\textbf{Blocks}} \\
    & & \ccell{\textbf{1}} & \ccell{\textbf{2}} & \ccell{\textbf{3}} & \ccell{\textbf{4}} & \ccell{\textbf{5}} & \ccell{\textbf{6}} & \ccell{\textbf{7}} & \ccell{\textbf{8}} & \ccell{\textbf{9}} & \ccell{\textbf{10}} & \ccell{\textbf{11}} & \ccell{\textbf{12}} & \ccell{\textbf{13}} & \ccell{\textbf{14}} & \ccell{\textbf{15}} & \ccell{\textbf{16}} & \ccell{\textbf{17}} & \ccell{\textbf{18}} & \ccell{\textbf{19}} & \ccell{\textbf{20}} & \ccell{\textbf{21}} & \ccell{\textbf{22}} \\
    \midrule
    \multirow{4}{*}{1} & right & tomorrow1 & date:8 & weather1 & afternoon1 & start1 & snow1: & :snow1 & - & temp2: & :temp2 & - & cold1 & - & danger1: & :danger1 & - & - & - &&&&\\
    & left & - & - & weather1 & afternoon1 & start1 & snow1: & :snow1 & - & temp2: & :temp2 & - & cold1 & - & -  & - & - & - & - &&&&\\
    & eye & - & - &- & - & - & - & EBf: & :EBf & - & - & - & - & - & - & EBf: & :EBf: & :EBf & - &&&&\\
    & mouth & - & - & - & - & - & - & - & - & - & Ci: & :Ci: & :Ci: & :Ci & - & - & - & Mo1: & :Mo1 &&&&\\ \midrule
    \multirow{4}{*}{2} & right & - & night1: & :night1 & start1: & :start1 & - & - & weekend1: & :weekend1 & - & date:10: & :date:10 & day1 & until1 & snow1: & :snow1 & - & temp2: & :temp2 & - & cold1 & danger1 \\
    & left & - & night1: & :night1 & start1: & :start1 & - & - & weekend1: & :weekend1 & - & - & - & day1 & until1 & snow1: & :snow1 & - & temp2: & :temp2 & - & cold1 & - \\
    & eye & - & - & - & - & - & - & - & - & - & - & - & - & - & - & - & EBf: & :EBf & - & - & - & - & - \\
    & mouth & Mmo: & :Mmo & - & - & Mmo: & :Mmo & Mmo: & :Mmo & - & Mmo: & :Mmo & - & - & - & - & - & - & - & Ci: & :Ci & - & - \\
    \bottomrule
\end{tabular}
}
\caption{Example blocks}
\label{tab:block_ex}
\end{table}

\renewcommand{\arraystretch}{1.4}

\begin{table}[p]
\centering
\small
\resizebox{\columnwidth}{!}{
\begin{tabular}{l *{6}{p{3cm}} |*{6}{p{3cm}}}
    \toprule
    \textbf{Order} & \multicolumn{12}{c}{\textbf{Grams}} \\
    & \multicolumn{6}{c|}{\textbf{Doc 1}} & \multicolumn{6}{c}{\textbf{Doc 2}} \\
    \cline{1-7} \cline{8-13}
    
    \multirow{4}{*}{\texttt{t1}} & eye\_EBf $\times 2$, & right\_tomorrow1, & right\_date:8, & right\_weather1, & right\_afternoon1, & right\_start1, &  eye\_EBf, & right\_night1, & right\_start1, & right\_weekend1, & right\_date:10, & right\_day1 \\
    & right\_snow1, & right\_temp2, & right\_cold1, & right\_danger1, & left\_weather1, & left\_afternoon1, &  right\_until1, & right\_snow1, & right\_temp2, & right\_cold1, & right\_danger1, & left\_night1, \\
    & left\_start1, & left\_snow1, & left\_temp2, & left\_cold1, & mouth\_Ci, & mouth\_Mo1 &  left\_start1, & left\_weekend1, & left\_day1, & left\_until1, & left\_snow1, & left\_temp2, \\
    &&&&&&& left\_cold1, & mouth\_Mmo $\times 4$, & mouth\_Ci &&& \\
    \cline{1-7} \cline{8-13}

    \multirow{6}{*}{\texttt{t2}} & (eye\_EBf & eye\_EBf), & (right\_tomorrow1 & right\_date:8), & (right\_date:8 & right\_weather1), & (right\_night1 & right\_start1), & (right\_start1 & right\_weekend1), & (right\_weekend1 & right\_date:10), \\
    & (right\_weather1 & right\_afternoon1), & (right\_afternoon1 & right\_start1), & (right\_start1 & right\_snow1), & (right\_date:10 & right\_day1), & (right\_day1 & right\_until1), & (right\_until1 & right\_snow1), \\
    & (right\_snow1 & right\_temp2), & (right\_temp2 & right\_cold1), & (right\_cold1 & right\_danger1), & (right\_snow1 & right\_temp2), & (right\_temp2 & right\_cold1), & (right\_cold1 & right\_danger1), \\
    & (left\_weather1 & left\_afternoon1), & (left\_afternoon1 & left\_start1), & (left\_start1 & left\_snow1), & (left\_night1 & left\_start1), & (left\_start1 & left\_weekend1), & (left\_weekend1 & left\_day1), \\
    & (left\_snow1 & left\_temp2), & (left\_temp2 & left\_cold1), & (mouth\_Ci & mouth\_Mo1) & (left\_day1 & left\_until1), & (left\_until1 & left\_snow1), & (left\_snow1 & left\_temp2) \\
    &&&&&&& (left\_temp2 & left\_cold1), & (mouth\_Mmo & mouth\_Mmo) $\times 3$, & (mouth\_Mmo & mouth\_Ci) \\
    \cline{1-7} \cline{8-13}

    \multirow{7}{*}{\texttt{c2}} & (left\_weather1 & right\_weather1), & (left\_afternoon1 & right\_afternoon1), & (left\_start1 & right\_start1), & (left\_night1 & right\_night1) $\times 4$, & (mouth\_Mmo & right\_night1), & (left\_night1 & mouth\_Mmo), \\
    & (left\_snow1 & right\_snow1) $\times 2$, & (eye\_EBf & right\_snow1), & (eye\_EBf & left\_snow1), & (left\_start1 & right\_start1) $\times 2$, & (mouth\_Mmo & right\_start1), & (left\_start1 & mouth\_Mmo), \\
    & (left\_temp2 & right\_temp2) $\times 2$, & (mouth\_Ci & right\_temp2), & (left\_temp2 & mouth\_Ci), & (left\_weekend1 & right\_weekend1) $\times 2$, & (mouth\_Mmo & right\_weekend1), & (left\_weekend1 & mouth\_Mmo), \\
    & (left\_cold1 & right\_cold1), & (mouth\_Ci & right\_cold1), & left\_cold1 & mouth\_Ci), & (mouth\_Mmo & right\_date:10), & (left\_day1 & right\_day1), & (left\_until1 & right\_until1), \\
    & (eye\_EBf & right\_danger1), & (eye\_EBf & mouth\_Mo1) & & & (left\_snow1 & right\_snow1) $\times 2$, & (eye\_EBf & right\_snow1), & (eye\_EBf & left\_snow1), \\
    &&&&&&& (left\_temp2 & right\_temp2) $\times 2$, & (mouth\_Ci & right\_temp2), & (left\_temp2 & mouth\_Ci), \\
    &&&&&&& (left\_cold1 & right\_cold1) & & & &\\
    \cline{1-7} \cline{8-13}
    
    \multirow{8}{*}{\texttt{t3}} & (right\_tomorrow1 & right\_date:8 & right\_weather1), & (right\_date:8 & right\_weather1 & right\_afternoon1), & (right\_night1 & right\_start1 & right\_weekend1), & (right\_start1 & right\_weekend1 & right\_date:10), \\

    & (right\_weather1 & right\_afternoon1 & right\_start1), & (right\_afternoon1 & right\_start1 & right\_snow1), & (right\_weekend1 & right\_date:10 & right\_day1), & (right\_date:10 & right\_day1 & right\_until1), \\

    & (right\_start1 & right\_snow1 & right\_temp2), & (right\_snow & right\_temp2 & right\_cold1), & (right\_day1 & right\_until1 & right\_snow1), & (right\_until1 & right\_snow1 & right\_temp2), \\

    & (right\_temp2 & right\_cold1 & right\_danger1), & (left\_weather1 & left\_afternoon & left\_start1), & (right\_snow1 & right\_temp2 & right\_cold1), & (right\_temp2 & right\_cold1 & right\_danger1), \\

    & (left\_start1 & left\_snow1 & left\_temp2), & left\_snow1 & left\_temp2 & left\_cold1) & (left\_night1 & left\_start1 & left\_weekend1), & (left\_start1 & left\_weekend1 & left\_day1),\\

    &&&&&&& (left\_weekend1 & left\_day1 & left\_until1), & (left\_day1 & left\_until1 & left\_snow1),\\
    &&&&&&& (left\_until1 & left\_snow1 & left\_temp2), & (left\_snow1 & left\_temp2 & left\_cold1), \\
    &&&&&&& (mouth\_Mmo &  mouth\_Mmo & mouth\_Mmo) $\times 2$, & (mouth\_Mmo & mouth\_Mmo & mouth\_Ci)\\
    \bottomrule
\end{tabular}
}
\caption{Example grams.}
\label{tab:grams_ex}
\end{table}

\end{landscape}

\end{document}